\documentclass[letterpaper, 10 pt, conference]{ieeeconf}  

\IEEEoverridecommandlockouts                              

\usepackage{cite, amsmath}
\usepackage{tabularx}
\usepackage{scalerel}
\usepackage{lipsum}
\usepackage{comment}
\usepackage{nicefrac}
\usepackage{dblfloatfix}
\usepackage{balance}

\usepackage{graphicx} 
\usepackage{amsmath} 
\usepackage{amssymb}  

\title{\LARGE \bf
Hybrid Volitional Control of a Robotic Transtibial Prosthesis using a Phase Variable Impedance Controller
}

\author{Ryan R. Posh$^{1}$,  Jonathan A. Tittle$^{1}$, David J. Kelly$^{1}$, James P. Schmiedeler$^{1}$, and Patrick M. Wensing$^{1}$
\thanks{*This work was funded by NSF grants DGE-1841556 \& CMMI-1943703.}
\thanks{$^{1}$All authors are with the Department of Aerospace and Mechanical Engineering, University of Notre Dame, Notre Dame, IN 46556, USA
        {\tt\small rposh@nd.edu}}%
}

\DeclareMathOperator{\atan}{atan2}
\begin{document}

\maketitle

\begin{abstract}

For robotic transtibial prosthesis control, the global kinematics of the tibia can be used to monitor the progression of the gait cycle and command smooth and continuous actuation. In this work, these global tibia kinematics are used to define a phase variable impedance controller (PVIC), which is then implemented as the nonvolitional base controller within a hybrid volitional control framework (PVI-HVC). The gait progression estimation and biomechanic performance of one able-bodied individual walking on a robotic ankle prosthesis via a bypass adapter are compared for three control schemes: a passive benchmark controller, PVIC, and PVI-HVC. The different actuation of each controller had a direct effect on the global tibia kinematics, but the average deviation between the estimated and ground truth gait percentage were 1.6\%, 1.8\%, and 2.1\%, respectively, for each controller. 
Both PVIC and PVI-HVC produced good agreement with able-bodied kinematic and kinetic references. As designed, PVI-HVC results were similar to those of PVIC when the user used low volitional intent, but yielded higher peak plantarflexion, peak torque, and peak power when the user commanded high volitional input in late stance. This additional torque and power also allowed the user to volitionally and continuously achieve activities beyond level walking, such as ascending ramps, avoiding obstacles, standing on tip-toes, and tapping the foot. 
In this way, PVI-HVC offers the kinetic and kinematic performance of the PVIC during level ground walking, along with the freedom to volitionally pursue alternative activities.



\end{abstract}

\section{Introduction}

\begin{figure}[b]
    \centering
    \includegraphics[width = 0.8\linewidth, trim={0.3cm 20cm 10.7cm 0cm},clip]{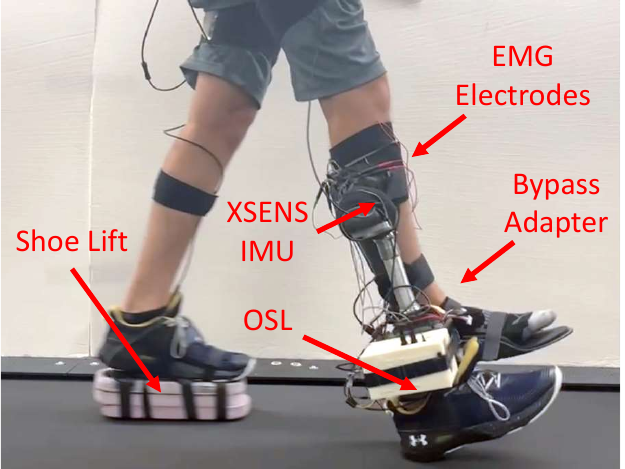}
    \caption{A single able-bodied user walked on the open-source leg (OSL) prosthesis via a bypass adapter. The user was outfitted with EMG electrodes and utilized an IMU on the tibia from an XSENS motion capture suit.} 
    \label{fig:Hardware}
\end{figure}

For individuals with transtibial amputation, robotic ankle prostheses could increase the mobility and quality of life compared to passive devices. While powered prosthetic hardware, as seen in Fig.~\ref{fig:Hardware}, has seen major advances in recent decades, improving control of these devices remains essential to returning individuals to full ability. For cyclic tasks, such as walking, controllers can predict the desired joint position, torque, or impedance by estimating the progression of the gait cycle. The most traditional approach is the use of a finite-state machine (FSM), which subdivides the gait cycle into discrete states that can be navigated based on sensor-based transition rules.
The discrete nature of FSM controllers leads to high repeatability, but also to an unnatural feel for users and the potential for state misclassification~\cite{posh2023finite}. 
%

To avoid discontinuous actuation, gait cycle progression can be continuously estimated by monitoring a phase variable. A phase variable, denoted herein as $\phi$, should progress monotonically from 0 to 100\% of the gait cycle and can be used to define a unique and continuous relationship between sensor signals and joint actuation. Phase variables that have been used include the foot center of pressure~\cite{gregg2013experimental, gregg2013towards} (only able to monitor gait progression in stance) and various hip and thigh kinematics~\cite{villarreal2014survey, cortino2022stair}. For transfemoral prostheses, thigh kinematics are preferred, as only sensors onboard the prosthesis are required. For transtibial prostheses, however, thigh-based phase variables require additional sensors on the user's body, increasing donning/doffing requirements. Therefore, a phase variable defined by the global kinematics of the tibia~\cite{holgate2009novel} 
is beneficial for transtibial prostheses since sensors onboard the prosthesis could measure it directly. 
Here, the global tibia angle is measured in the sagittal plane relative to the gravity-fixed vertical.
\begin{figure*}[t!]
    \centering
    \includegraphics[width = 0.9\linewidth, trim={0cm 22.4cm 0cm 0cm},clip]{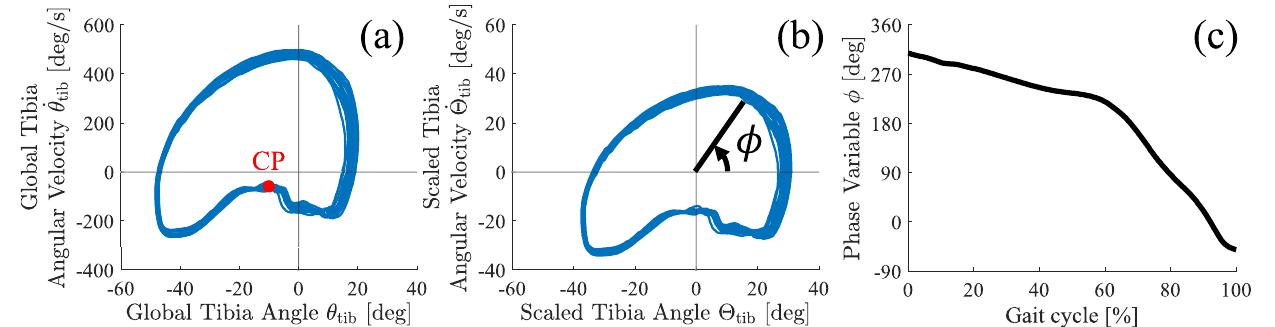}
    \vspace{-0.1cm}
    \caption{The uncalibrated phase portrait with critical point CP (a) must be scaled and shifted (b) to produce a monotonic relationship between phase variable and gait percentage (c)~\cite{posh2023IROS}.} 
    \label{fig:Phase_Calibration}
    \vspace{-0.5cm}
\end{figure*}

For continuous impedance control, a desired stiffness, damping, and equilibrium angle must be predefined as functions of gait percentage. At each percentage of gait, multiple combinations of these impedance parameters could yield the same joint torque. The ideal impedance trajectories to command remain an open topic of research. Some have defined these trajectories via Parameter Dependent Lyapunov Functions~\cite{mohammadi2019variable}, with optimization techniques~\cite{hu2016analysis, best2023data}, or with heuristic linear definitions~\cite{hong2019consolidated}, while others have looked to empirically measure these parameters on the human body~\cite{lee2016summary}. Some have worked to unify these findings~\cite{kumar2020impedance}, though consensus has not been fully reached. 

While a smooth and continuous phase-based control approach can be applied to ambulation on level ground~\cite{gregg2013experimental}, ramps~\cite{best2021phase}, and stairs~\cite{cortino2022stair}, non-cyclic tasks such as standing on tip-toes, tapping the foot, or stepping side-to-side require the incorporation of direct user intent. Direct myoelectric control (DMC) uses sensing of the lower leg muscle activity to provide users a direct sense of control and embodiment with the prosthetic ankle. DMC has shown promise in weight-bearing activities, including walking~\cite{huang2015locomotor, posh2023finite, hargrove2013robotic}, standing on tip-toes~\cite{shah2022design,posh2023finite}, and perturbation reactions~\cite{fleming2019proportional2}.

Hybrid Volitional Control (HVC) is a framework that has been shown to effectively combine standard walking controllers (such as FSM impedance control) with direct volitional controllers (such as DMC)~\cite{hoover2012stair, wang2013proportional, chen2015combining, posh2021hybrid, posh2023finite}. 
When applied to the full gait cycle, HVC allows users both to reliably walk on level ground with or without direct user intent and to perform non-cyclic activities by appropriately activating the instrumented muscles. 
Discrete transition-based controllers, however, may not be ideal as the nonvolitional base controller within HVC due to potential interplay with the volitional component at transitions~\cite{posh2023finite}. HVC has been combined with phase-variable control~\cite{posh2021hybrid}, though only in simulation. 

Therefore, this work seeks to introduce and assess 1) the first transtibial phase-variable impedance controller based on the global tibia kinematics, and 2) implementation of a hybrid volitional controller using a phase variable impedance controller (PVI-HVC). PVI-HVC appears to be a promising way for individuals who have experienced transtibial amputation to smoothly achieve both cyclic and non-cyclic tasks of daily living.

\section{Methods}
\subsection{Hardware}
\label{sec:hardware}

Each controller (passive, PVIC, and PVI-HVC) was implemented on the version 1 Open-Source Leg (OSL) \cite{azocar2018design}, which has a maximum range of motion of $\pm$15 degrees. All sensors were wired to a Raspberry Pi microcomputer, which received wireless commands from a remote laptop. For sensing, the OSL interfaced with two pressure sensors (SEN-08685; SparkFun Electronics), one each at the heel and toe, laid between two layers of polyurethane rubber compound (Vytaflex-60) and held in place by plastic 3D-printed contact plates. Each unit was secured to the bottom of the cosmetic footshell (LP Variflex\textregistered  foot), which was itself enveloped by a shoe. The user also donned the XSENS MVN Link motion capture suit for real-time measurement of the global tibia angle and angular velocity. Other data from the suit was used for lower body kinematic analysis offline. The XSENS suit communicated to the Raspberry Pi wirelessly via a private Wi-Fi router. 
A single able-bodied user walked on the prosthesis via a bypass adapter (Fig.~\ref{fig:Hardware}). A contralateral shoe lift was donned to accommodate leg length difference from the adapter. The user was outfitted with two surface electromyography (sEMG) amplifiers (SX230FW; Biometrics Ltd.), one each for the gastrocnmeius (GAS) and tibialis anterior (TA). To avoid irritation or discomfort, thin ($\sim$1.25mm thickness), compliant medical-grade Ag/AgCl disk electrodes (TE/K50430-001, Technomed USA) interfaced with the user's skin. Hair was removed and the skin cleaned as necessary. Electrodes were placed to align with the primary muscle fiber directions and were wrapped with soft velcro straps for security.

\subsection{Experimental Procedure}
After all calibrations, the user performed four 30-second steady-state walking trials on a level treadmill at 0.8 m/s while holding a handrail. 
Trial 1 used a benchmark passive controller, Trial 2 used PVIC, Trial 3 used PVI-HVC with low volitional user intent, and Trial 4 used PVI-HVC with high intent.

\subsection{Benchmark Passive Controller}
For calibration and a point of reference, a passive controller was implemented to mimic the behavior of a passive ankle-foot prosthesis. A constant impedance command was sent to the OSL with an ankle equilibrium angle set point of 0 degrees, a stiffness gain of 0.09 Nm/deg/kg, and a damping gain of 0.075 Nms/deg/kg.

\begin{figure*}[t!]
    \centering
    \includegraphics[width = 0.9\linewidth, trim={0cm 22.5cm 0cm 0cm},clip]{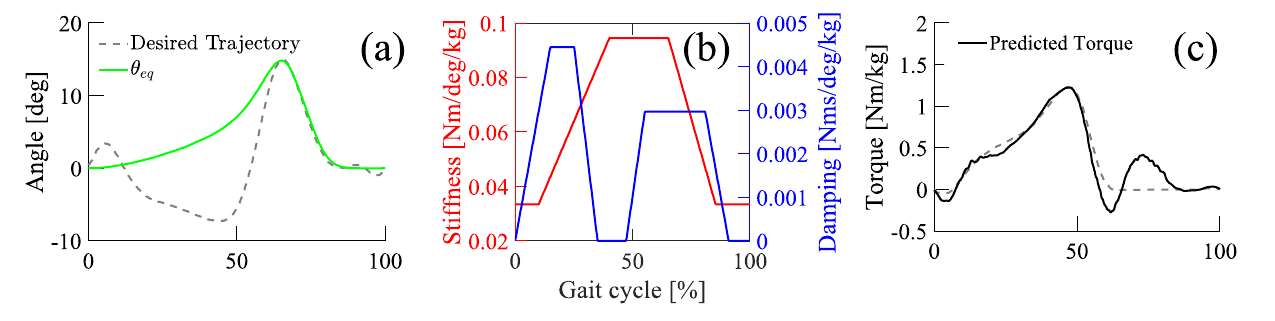}
    \vspace{-0.3cm}
    \caption{The ankle equilibrium angle (a), ankle stiffness, and ankle damping (b) are continuous with respect to gait percentage. These parameters define the impedance model to achieve the predicted torque (c). Desired trajectories for ankle angle and torque (dashed lines) are adapted from~\cite{winter1983biomechanical}.}
    \label{fig:IMP}
    \vspace{-0.5cm}
\end{figure*}
\subsection{Phase Variable Impedance Controller}
\label{sec:PVIC}

As in~\cite{holgate2009novel}, PVIC uses a global tibia-based phase variable. Calibration first established the phase relationship between tibia kinematics and gait percentage, and then a predefined impedance model was applied to achieve continuous impedance control.

\subsubsection{Tibia-based Phase Variable}
A phase portrait can be defined, as in Fig.~\ref{fig:Phase_Calibration}a, with the tibia angular displacement ($\theta$) on the horizontal axis and the tibia angular velocity ($\dot{\theta}$) on the vertical axis. In this phase portrait, heel strike occurs in the bottom right quadrant, where $\theta_{\rm tib}$ is positive and $\dot{\theta}_{\rm tib}$ is negative, and $\phi$ progresses clockwise from there through the gait cycle. 
To ensure a monotonic relationship between $\phi$ and gait percentage, $\theta_{\rm tib}$ and $\dot{\theta}_{\rm tib}$ are scaled and shifted with
 calibration constants $x_{0}$, $y_{0}$, and $k$ as in Fig.~\ref{fig:Phase_Calibration}b~\cite{posh2023IROS}.
\begin{eqnarray}
   \Theta_{\rm tib}(t) &=& \theta_{\rm tib}(t) - x_{0} \label{eq:Theta}\quad \quad \textrm{and}\\
   \dot{\Theta}_{\rm tib}(t) &=& k(\dot{\theta}_{\rm tib}(t) - y_{0})
   \label{eq:Theta_dot}
\end{eqnarray}
\noindent are then used to compute the final phase variable 
\begin{equation}
    \phi(t) = \atan(\dot{\Theta}_{\rm tib}(t), \Theta_{\rm tib}(t)).
    \label{eq:phi}
\end{equation}
\noindent As the gait cycle progresses from 0 to 100\%, $\phi$ (starting between 270 and 360 degrees) should monotonically decrease clockwise around the phase portrait until reaching the next heel strike (Fig.~\ref{fig:Phase_Calibration}c).

\subsubsection{Calibration} 
To determine the calibration parameters in Eqs.~\ref{eq:Theta} and ~\ref{eq:Theta_dot}, the user walked at a steady-state speed of 0.8m/s for several strides (10-20) with the benchmark passive controller engaged.
During these strides, the global tibia angle and angular velocity were recorded, along with pressure sensor data. Heel strike, identified when the heel pressure exceeds a predefined threshold, establishes a ground truth gait percentage by normalizing time between each heel strike. Critical Point Centering (CPC) calibration from~\cite{posh2023IROS} shifts the phase portrait based on its critical point where the global tibia angular velocity during stance is maximum (negative value in Fig.~\ref{fig:Phase_Calibration}a). 
The parameter $x_0$ is chosen such that the average critical point is located on the y-axis or
$x_0 = \theta_{\rm tib}(\max(\dot{\theta}_{\rm tib_{stance}}))$.
The parameter $y_0$ is chosen to vertically center the phase portrait about the horizontal axis, or $y_{0} = (\overline{\dot{\theta}}_{\rm tib} + \underline{\dot{\theta}}_{\rm tib})/2$,
and $k$ is chosen to scale the phase portrait such that the maxima/minima of $\theta_{\rm tib}$ and $\dot{\theta}_{\rm tib}$ are approximately equal, or
\begin{equation*}
   k = \frac{|\overline{\theta}_{\rm tib} - \underline{\theta}_{\rm tib}|}{|\overline{\dot{\theta}}_{\rm tib} - \underline{\dot{\theta}}_{\rm tib}|}.
\end{equation*}
Successful calibration yields a monotonic relationship between ground truth gait percentage and progression of phase variable $\phi$, as in Fig.~\ref{fig:Phase_Calibration}c. This relationship can then be used to transform a desired impedance model expressed as a function of gait percentage to one expressed as a function of phase variable progression.
\begin{figure}[t!]
    \centering
    \includegraphics[width = 0.90\linewidth, trim={0cm 17.5cm 0cm 0cm},clip]{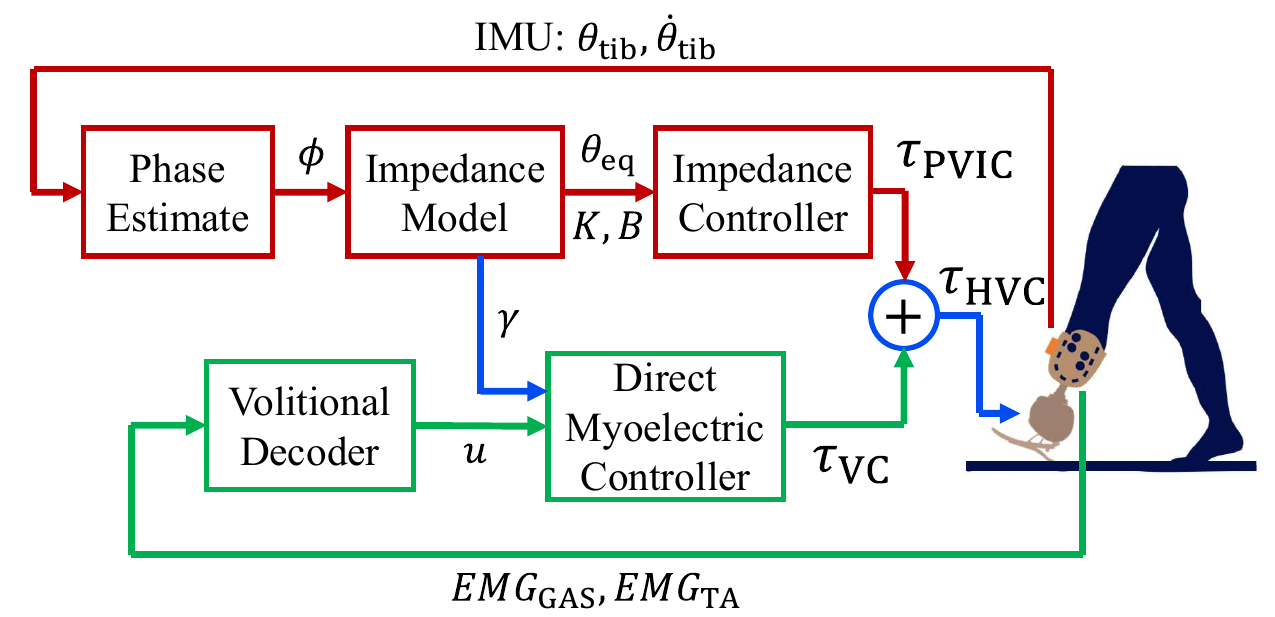}
    \vspace{-0.1cm}
    \caption{Hybrid volitional control (HVC, blue) is an additive combination of torque from phase variable impedance control (PVIC, red) and volitional control (VC, green).  } 
    \label{fig:Block}
    \vspace{-0.5cm}
\end{figure}
\begin{figure*}[t!]
    \centering
    \includegraphics[width = 0.9\linewidth, trim={0cm 21.5cm 0cm 0cm},clip]{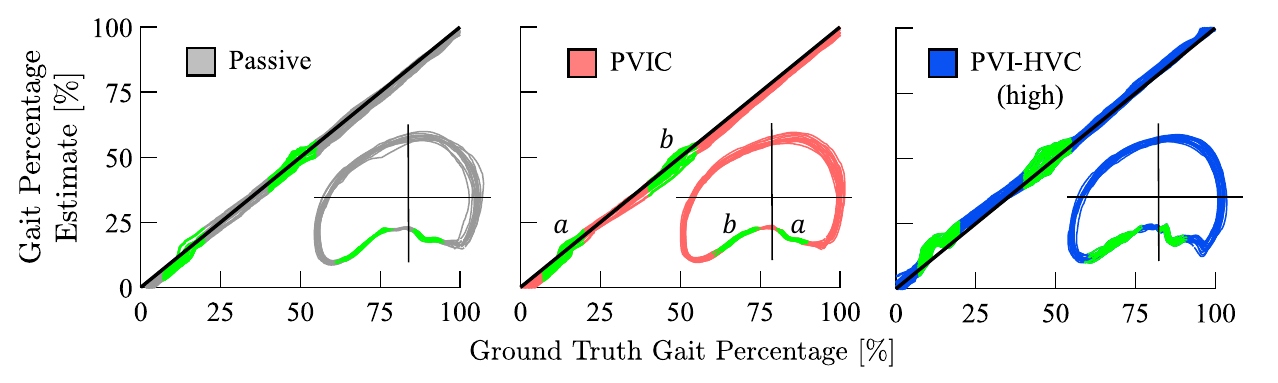}
    \vspace{-0.3cm}
    \caption{Measurement of the scaled phase portrait leads to the estimation of the gait percentage for passive control (grey), PVIC (red), and PVI-HVC with high intent (darker blue). Highlighted in green are regions of increased prediction error with corresponding locations \textit{a} and \textit{b}.} 
    \label{fig:Estimation}
    \vspace{-0.5cm}
\end{figure*}

\subsubsection{Impedance Model and Control} 
\label{sec:IMPModel}
This work uses a heuristically defined impedance model that prioritizes a desired torque trajectory in stance and a desired position trajectory in swing, similar to the knee-ankle controller in~\cite{best2023data}. 
It consists of desired trajectories for the ankle equilibrium angle, stiffness gain, and damping gain as functions of gait percentage, shown in Fig.~\ref{fig:IMP}. The equilibrium angle trajectory (Fig.~\ref{fig:IMP}a) during swing matches a scaled version of the desired able-bodied swing phase kinematics, beginning at the point of maximum plantarflexion. The able-bodied reference was adapted from~\cite{winter1983biomechanical}, but the maximum plantarflexion angle was scaled down from approximately 20 degrees to match the OSL's plantarflexion limit of 15 degrees. The equilibrium angle trajectory during stance was defined such that the difference between the desired able-bodied kinematics and the equilibrium angle could be linearly scaled to match the desired ankle torque. This scaling was achieved via a simply defined stiffness trajectory that linearly increased during mid-stance as in~\cite{lee2016summary}, remained constant during push-off, and linearly decreased in late swing to ensure that the values at 0\% and 100\% of gait were equal for continuity across heel strike (Fig.~\ref{fig:IMP}b). The heuristically defined damping trajectory consists of two peaks, as observed in~\cite{lee2016summary}, one that assists with shock absorption in early stance and the other that dampens the position tracking in swing to reduce any oscillations related to overshoot (Fig.~\ref{fig:IMP}b).

As outlined in Fig.~\ref{fig:Block}, measuring the progression of the phase variable enables a continuous estimate of gait percentage such that the desired stiffness, damping, and equilibrium angle can be applied. The relationship between this impedance model and the output torque is
\newcommand\Tau{\scalerel*{\tau}{T}}
\begin{equation}
    \Tau_{\rm PVIC} = -K(\theta_{\rm ankle} - \theta_{\rm eq}) - B\dot{\theta}_{\rm ankle},
\end{equation}

\noindent where the stiffness $K$, damping $B$, and ankle equilibrium angle $\theta_{\rm eq}$ are predefined functions of gait percentage, and, after calibration, also functions of the phase variable $\phi$. Using able-bodied kinematics for reference, the expected torque from this impedance model is shown in Fig.~\ref{fig:IMP}c.

\subsection{Phase Variable Impedance Hybrid Volitional Controller}



As outlined in~\cite{posh2021hybrid}, HVC sums the torque contributions from two components: a non-volitional base component (NVBC) and a volitional component. With PVIC (Sec.~\ref{sec:PVIC}) serving as the NVBC, the output torque is
\begin{equation}
    \Tau_{\rm PVI-HVC} = \Tau_{\rm PVIC} + \Tau_{\rm VC}.
\end{equation}
\noindent In this way, users can rely on the NVBC to walk naturally and reliably on level ground without the need to use any volitional input, but can also volitionally contribute at any time to achieve non-standard activities, navigate obstacles or uneven terrain, or increase the torque and power of the NVBC when walking, if desired.

The volitional component follows~\cite{posh2023finite} and converts EMG signals from the GAS and TA to a single volitional input variable,
\begin{eqnarray}
u = 
\begin{cases}
    \sqrt{u_p^2 + u_d^2} \cdot (\frac{m - m_0}{m_{\rm GAS} - m_0}) &\text{if $m \geq m_0$} \\
    -\sqrt{u_p^2 + u_d^2} \cdot (\frac{m - m_0}{m_{\rm TA} - m_0}) &\text{if $m < m_0$}
\end{cases}\,, \label{eq:u}\\
u_p = \rm EMG_{\rm GAS}/MVA_{\rm GAS}, \\
u_d = \rm EMG_{\rm TA}/MVA_{\rm TA}, \\
m = u_p/u_d, \\
m_0 = {\rm atan}(\nicefrac{{\rm tan}(m_{\rm GAS})+{\rm tan}(m_{\rm TA})}{2}),
\end{eqnarray}
\noindent where $u_p$ ($p$ for plantarflexion) and $u_d$ ($d$ for dorsiflexion) are the GAS and TA sEMG inputs, normalized by the maximum voluntary activation (MVA) level for each muscle, $m$ is the real-time co-contraction slope, and $m_{\rm GAS}$ and $m_{\rm TA}$ are the average co-contraction slopes measured when the user attempts to strictly plantar- or dorsiflex, respectively. The MVA$_{\rm GAS}$, MVA$_{\rm TA}$, $m_{\rm GAS}$, $m_{\rm TA}$, and bisector $m_0$ are all parameters found during volitional calibration and are input to the volitional decoder in Fig.~\ref{fig:Block}, defined by Eq.~\ref{eq:u}. The $\sqrt{u_p^2 + u_d^2}$ term is constrained to never exceed a value of 1, and the value of $m$ is constrained to never be larger than $m_{\rm GAS}$ or smaller than $m_{\rm TA}$. Therefore, the volitional input $u$ ranges from 1 for maximum plantarflexion intent to -1 for maximum dorsiflexion intent. 

To calibrate the volitional component of PVI-HVC, the user performed 5 maximum voluntary isometric contractions for plantarflexion (maximum defining MVA$_{\rm GAS}$) and for dorsiflexion (maximum defining MVA$_{\rm TA}$). Co-contraction parameters were found via a dynamic calibration, where the user walked on the treadmill for 30 seconds using the benchmark passive controller. The parameter $m_{\rm GAS}$ was calculated as the average slope $m$ for conditions when $u_p > 0.5$ and $u_d < 0.5$, and $m_{\rm TA}$ was calculated as the average slope when $u_p < 0.5$ and $u_d > 0.5$.
Ultimately, the torque from the volitional component is 
\begin{eqnarray}
    \Tau_{\rm VC} = 
    -K \gamma u ,\\
    \gamma = \theta_{\rm max} - |\theta_{\rm eq}|,
\end{eqnarray}
\noindent where $\theta_{\rm max}$ is the maximum plantarflexion angle of the prosthesis (15 degrees) and $K$ and $\theta_{\rm eq}$ are the stiffness and equilibrium angle, respectively, defined by the impedance model in Sec~\ref{sec:IMPModel}. When combined with the PVIC, this control acts to volitionally modulate the equilibrium angle up to the maximum range of the prosthesis. 

\subsection{Data Processing}

The OSL's motor encoder measured the motor position, which was mapped to the ankle angle via the OSL's four-bar mechanism kinematics~\cite{azocar2018design}. 
The sensed motor current was converted to motor torque and combined with the four-bar kinematics to estimate ankle torque. The prosthetic ankle joint power was computed from these quantities.
The sEMG sensors feature a band-pass filter between 20 and 460 Hz and 1000x amplification.
The system operated at 220 Hz (limited by the XSENS suit communication), with the rectified sEMG signals, pressure sensor signals, and global tibia velocity filtered in the time domain with moving average filters defined by 300 ms, 60 ms, and 70 ms windows, respectively. By smoothing the velocity, the phase portrait is slightly tilted due to a small delay between global tibia angle and angular velocity. As mentioned in~\cite{holgate2009novel}, the physical meanings of the phase portrait axes are unimportant in steady-state activities as long as the inputs uniquely correspond to the desired gait percentage.
The data were divided into strides based on heel strike occurrences. 
No strides were removed as outliers.

\section{Results \& Discussion}

%
\begin{figure*}[t!]
    \centering
    \includegraphics[width = \linewidth, trim={0cm 10cm 0cm 0cm},clip]{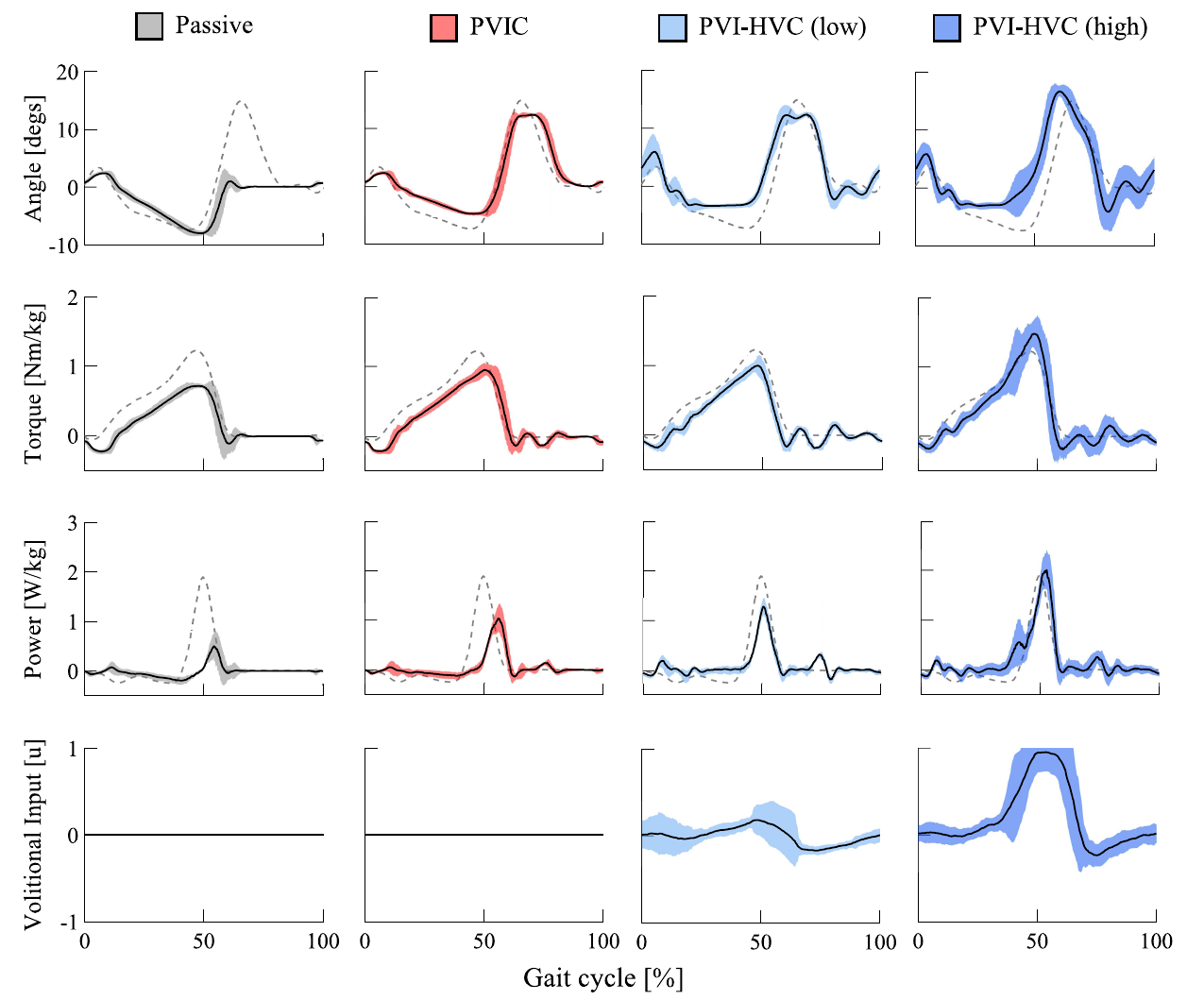}
    \vspace{-0.5cm}
    \caption{After treadmill walking, the mean prosthesis ankle angle, normalized torque, normalized power, and volitional input versus gait cycle percentage were measured for passive control (grey), PVIC (red), PVI-HVC with low intent (lighter blue), and PVI-HVC with high intent (darker blue). Each plot shows the average trajectory $\pm$ 2 standard deviations. Desired trajectories (dashed grey) were adapted from able-bodied walking~\cite{winter1983biomechanical}.}
    \label{fig:Comparison}
    \vspace{-0.3cm}
\end{figure*}

\subsection{Steady State Phase Estimation}

Figure~\ref{fig:Estimation} shows the scaled phase portrait and gait percentage estimation performance for all of the recorded strides when walking with passive control, PVIC, and PVI-HVC with high intent. 
As more actuation is introduced into the control scheme, the tibia kinematics, and therefore the phase variable and estimated gait percentage, are increasingly affected at particular locations. The maximum differences between percentage estimate and ground truth gait percentage for all strides were 4.8$\%$ for passive (occurring at 55.6$\%$ ground truth gait percentage), 4.7$\%$ for PVIC (at 8.1$\%$ ground truth), and 8.3$\%$ (at 44.8$\%$ ground truth) for PVI-HVC. These larger deviations around 10$\%$ and 50$\%$ ground truth percentage, highlighted in Fig.~\ref{fig:Estimation}, correspond to $\phi$ values of around 290 and 235 degrees in Fig.~\ref{fig:Phase_Calibration}c. At these locations, a near zero slope for the phase variable-gait percentage relationship leads to small changes in $\phi$ corresponding to relatively large changes in estimated gait percentage.
Despite these localized errors, the average differences between percentage estimate and ground truth gait percentage across the full stride were 1.6$\%$, 1.8$\%$, and 2.1$\%$ for passive, PVIC, and PVI-HVC, respectively. 


\subsection{Biomechanic Performance}

Figure~\ref{fig:Comparison} shows results of the 30-second steady-state treadmill walking trials with the 3 different controllers: passive control (18 strides), PVIC (16 strides), and PVI-HVC shown with both low (16 strides) and high (17 strides) user intent. 
Each plot is the average trajectory $\pm$ 2 standard deviations of the recorded strides. Desired able-bodied trajectories are shown for reference.
The expected power trajectory was determined by multiplying the desired torque and the time derivative of the desired angle.

\subsubsection{Ankle Kinematics}
The ankle angle trajectory generated by the passive controller shows natural dorsiflexion in stance due to loading conditions, but no significant plantarflexion related to toe-off, resulting in a root-mean-squared error (RMSE) from the desired trajectory of 5.1 degrees. With an equilibrium angle of 0 degrees, the ankle remains neutral in the absence of ground reaction forces, such as in swing, and the constant damping gain reduces oscillations during the stance-to-swing transition. PVIC generated a smooth, continuous, and highly repeatable ankle angle trajectory, with an RMSE of 2.2 degrees from desired, and an average peak plantarflexion angle of 12.5 degrees. Slightly less dorsiflexion than desired is observed in stance, which is likely due to the atypical loading conditions associated with walking with a bypass adapter and holding the treadmill handrail. 

As expected, PVI-HVC with low intent showed similar performance to PVIC (RMSE of 2.9 degrees). There were some increases in oscillation and variance around heel strike and less dorsiflexion during stance, all attributed to the baseline volitional input, as observed in Fig.~\ref{fig:Comparison}, from the unintentional muscle activation involved in walking with low intent. PVI-HVC with high user intent similarly generated less dorsiflexion in early stance due to this baseline volitional input, but the user was able to volitionally command a large plantarflexion intent between 40 and 75$\%$ of gait to exceed the desired peak plantarflexion angle on average. PVI-HVC showed the highest stride-to-stride variation due to the variability of human input. Overall, the ankle angle RMSE of PVI-HVC with high intent was 4.3 degrees.

\subsubsection{Ankle Kinetics}

Despite an inability to track the desired ankle kinematics, the passive controller produces a smooth and natural torque trajectory that is qualitatively similar to, but quantitatively less than the desired torque trajectory, with an RMSE of 0.25 Nm/kg. Peak power, however, reached only 30\% of the desired on average. Again, PVIC produced a smooth and repeatable torque trajectory that was very close to the desired (RMSE of 0.18 Nm/kg). A smaller plantarflexion moment was observed in stance, which is consistent with a reduction in the dorsiflexion angle related to loading. The max power generated was 1.1 W/kg on average. As expected, PVI-HVC generated very similar torque (RMSE of 0.17 Nm/kg) and power (peak of 1.3 W/kg) trajectories to PVIC when volitional input remained low.
PVI-HVC with high intent produced torque and power trajectories very similar to the desired, with a maximum plantarflexion moment of 1.6 Nm/kg and a maximum power of 2.0 W/kg. PVC-HVC with high intent produced the least kinetic error from desired, with RMS torque and power errors of 0.14 Nm/kg and 0.25 W/kg, respectively.

\subsection{Volitional Activities}

\begin{figure}[t!]
    \centering
    \includegraphics[width = \linewidth, trim={0.3cm 22cm 10.7cm 0cm},clip]{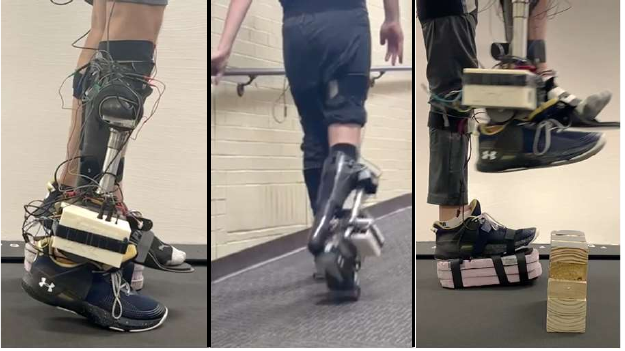}
    \vspace{-0.5cm}
    \caption{With volitional input, users can stand on tip-toes, ascend ramps, avoid obstacles, and perform other activities (see supplemental video).} 
    \vspace{-0.5cm}
    \label{fig:nonstandard}
\end{figure}

By contributing volitional input when desired, the user was able to achieve activities beyond level ground walking, including standing on tip-toes, tapping the foot, ascending a ramp, and avoiding obstacles in swing (Fig.~\ref{fig:nonstandard}, and supplemental video). This is a major advantage of the PVI-HVC framework. When the tibia segment remains relatively static, the controller approximates a DMC implementation.

\subsection{Controller Strengths \& Study Limitations}

To the authors' knowledge, this is the first implementation of a variable impedance controller for an ankle prosthesis that makes use of a tibia-based phase variable. This PVIC not only offers high accuracy gait estimation, but also matches qualitatively and quantitatively very well with intact ankle kinematics and kinetics for walking on level ground. Adding myoelectric volitional input to the system in PVI-HVC extends this controller into one that can also achieve activities of daily living beyond level walking.
Unlike finite-state machine HVC~\cite{posh2023finite}, this continuous controller did not exhibit unexpected behaviors from the interactions between the PVIC and the volitional component.



A limitation of this study is the range of motion of the OSL, which only allows for proper comparisons to slow able-bodied walking. By modifying the impedance model, this framework can easily be extended to other hardware platforms that allow for a wider range of motion. Additionally, the data in this study come from only one able-bodied user, and the effects on biomechanic performance of walking on the prosthesis via a bypass adapter are non-trivial. Future work will assess the performance of these controllers with multiple subjects with transtibial amputation. 

\section{Conclusion}
This paper presents the first phase-variable impedance controller (PVIC) based on the global tibia kinematics and employs this controller in a hybrid volitional control framework (PVI-HVC) to achieve both cyclic and non-cyclic tasks. In comparisons among a baseline passive controller, PVIC implemented alone, and PVI-HVC, the different actuation associated with each had a direct effect on the gait percentage estimation 
at certain gait percentages.
On average, though, the gait percentage predictions when using all three all had an average accuracy of approximately 2\%. For level ground walking, PVIC approximated the kinematic and kinetic performance of an intact ankle very well.
PVI-HVC with low user volitional input retained the performance of PVIC, and adding high volitional input increased the degree of plantarflexion, peak torque, and peak power when walking, resulting in the best agreement to able-bodied torque and power references. Beyond level walking, PVI-HVC also makes it possible to achieve activities such as slope walking, obstacle avoidance, and tip-toe standing. 

\balance









\pagebreak
\bibliographystyle{IEEEtran}
\bibliography{References}

\end{document}